\documentclass{IOS-Book-Article}

\usepackage{mathptmx}
\usepackage{soul}\setuldepth{article}
\usepackage[T1]{fontenc}
\usepackage[utf8]{inputenc}
\usepackage{listings}
\usepackage{xcolor}
\usepackage{upquote}
\usepackage{graphicx}
\usepackage{float}
\usepackage{booktabs}
\usepackage{subcaption}

\colorlet{punct}{red!60!black}
\definecolor{background}{gray}{0.98}
\definecolor{delim}{RGB}{20,105,176}
\colorlet{numb}{magenta!60!black}

\lstdefinelanguage{json}{
    basicstyle=\footnotesize\ttfamily,
    columns=fullflexible,
    keepspaces=true,
    numbers=left,
    numberstyle=\scriptsize\color{black!25},
    stepnumber=1,
    numbersep=4pt,
    showstringspaces=false,
    breaklines=true,
    frame=single,
    framerule=0pt,
    framesep=2pt,
    rulecolor=\color{black},
    backgroundcolor=\color{background},
    xleftmargin=2pt, xrightmargin=2pt,
    aboveskip=6pt, belowskip=6pt,
    firstnumber=0,
    literate=
     *{0}{{{\color{numb}0}}}{1}
      {1}{{{\color{numb}1}}}{1}
      {2}{{{\color{numb}2}}}{1}
      {3}{{{\color{numb}3}}}{1}
      {4}{{{\color{numb}4}}}{1}
      {5}{{{\color{numb}5}}}{1}
      {6}{{{\color{numb}6}}}{1}
      {7}{{{\color{numb}7}}}{1}
      {8}{{{\color{numb}8}}}{1}
      {9}{{{\color{numb}9}}}{1}
      {:}{{{\color{punct}{:}}}}{1}
      {,}{{{\color{punct}{,}}}}{1}
      {\{}{{{\color{delim}{\{}}}}{1}
      {\}}{{{\color{delim}{\}}}}}{1}
      {[}{{{\color{delim}{[}}}}{1}
      {]}{{{\color{delim}{]}}}}{1}
      {Ä}{{\"A}}1 {Ö}{{\"O}}1 {Ü}{{\"U}}1
      {ä}{{\"a}}1 {ö}{{\"o}}1 {ü}{{\"u}}1
      {ß}{{\ss}}1 {§}{{\S}}1,
}

\hypersetup{
  colorlinks=true,
  allcolors=blue
}

\def\hb{\hbox to 11.5 cm{}}

\begin{document}

\pagestyle{headings}
\def\thepage{}
\begin{frontmatter}              

\title{Segmentation and Processing of German Court Decisions from Open Legal Data}

\markboth{}{}

\author[A,B]{\fnms{Harshil} \snm{Darji}\orcid{0000-0002-8055-1376}%
\thanks{Corresponding Author: Harshil Darji, Harshil.Darji@HTW-Berlin.de.}},
\author[A]{\fnms{Martin} \snm{Heckelmann}\orcid{0000-0003-1696-6849}},
\author[A]{\fnms{Christina} \snm{Kratsch}\orcid{0000-0003-0565-1112}}
and
\author[B]{\fnms{Gerard} \snm{de Melo}\orcid{0000-0002-2930-2059}}

\runningauthor{H. Darji et al.}
\address[A]{Hochschule für Technik und Wirtschaft Berlin, Germany}
\address[B]{Hasso-Plattner Institute / University of Potsdam, Germany}

\begin{abstract}
The availability of structured legal data is important for advancing Natural Language Processing (NLP) techniques for the German legal system. One of the most widely used datasets, Open Legal Data, provides a large-scale collection of German court decisions. While the metadata in this raw dataset is consistently structured, the decision texts themselves are inconsistently formatted and often lack clearly marked sections. Reliable separation of these sections is important not only for rhetorical role classification but also for downstream tasks such as retrieval and citation analysis. In this work, we introduce a cleaned and sectioned dataset of \textbf{251,038} German court decisions derived from the official Open Legal Data dataset. We systematically separated three important sections in German court decisions, namely \textit{Tenor} (operative part of the decision), \textit{Tatbestand} (facts of the case), and \textit{Entscheidungsgründe} (judicial reasoning), which are often inconsistently represented in the original dataset. To ensure the reliability of our extraction process, we used Cochran's formula with a \textbf{95\%} confidence level and a \textbf{5\%} margin of error to draw a statistically representative random sample of \textbf{384} cases, and manually verified that all three sections were correctly identified. We also extracted the \textit{Rechtsmittelbelehrung} (appeal notice) as a separate field, since it is a procedural instruction and not part of the decision itself. The resulting corpus is publicly available in the JSONL format, making it an accessible resource for further research on the German legal system.
\end{abstract}

\end{frontmatter}
\markboth{}{}

\section{Introduction}
The increasing volume of legal data and growing digitization of court decisions in Germany has created new opportunities to advance NLP in the legal domain. Large-scale initiatives, such as \textit{Open Legal Data}~\cite{r1} and \textit{GerDaLIR}~\cite{r2}, provide access to thousands of court decisions across different courts and years. However, the usability of such resources in Legal NLP is affected by inconsistent formatting, missing structural markers, and varying HTML structures \cite{r3, r4}.

The structure of German court decisions follows a consistent legal writing format, where \textit{Tenor} provides the operative part of the judgment, \textit{Tatbestand} summarizes facts and procedural history, and \textit{Entscheidungsgründe} outline the reasoning. For downstream tasks such as legal case retrieval \cite{r5}, citation analysis, and rhetorical role classification \cite{r6}, a reliable separation of these sections is a prerequisite \cite{r7}. A lack of clear segmentation can directly affect retrieval, where overlapping reasoning with factual history can lead to false matches, and citation analysis, where the significance of a cited statute depends on whether it appears in the reasoning or the operative part. In Retrieval-Augmented Generation (RAG) systems, section-aware chunking can improve interpretability and prevent the model from mixing argumentative and operative content.

Prior research on German court decisions has also highlighted the importance of structured and machine-readable court decisions for retrieval and classification \cite{r8, r9, r10}, but existing datasets still lack consistent section boundaries. In this paper, we introduce a dataset of 251,038 German court decisions derived from the official Open Legal Data data dump. Using a rule-based extraction approach with regular expressions, we separated the three decision sections and also identified statutory and case references to make interlinks within the corpus readily available. To ensure reliability, we used Cochran’s formula \cite{r11} with a 95\% confidence level and 5\% margin of error, drawing a random sample of 384 cases that we manually verified. In this sample, \textbf{97.40\%} ($\pm 1.59\%$) of the cases were correctly processed, confirming the reliability of the extraction procedure. The dataset is publicly available in JSONL format via Hugging Face datasets\footnote{harshildarji/openlegaldata. \url{https://huggingface.co/datasets/harshildarji/openlegaldata}}.

\section{Related Work}

The \textit{Open Legal Data platform} introduced by Ostendorff et al. \cite{r1} created a large corpus of publicly accessible German court decisions, with metadata including court identifiers and dates. However, while there is a consistent structure for metadata, the decision texts are stored as heterogeneous HTML and lack standardized markup. Several subsequent corpora have since been derived from this source. Glaser et al. \cite{r12} created a collection of approximately 100,000 German court rulings to evaluate summarization methods, segmenting texts into units suitable for training summarization models, though without enforcing legal section boundaries. Wrzalik et al. \cite{r2} introduced \textit{GerDaLIR}, an information retrieval benchmark linking 123,000 query passages with 131,000 documents based on citations between decisions. Darji et al. \cite{r3} explored semantic similarity between German court decisions and statutory provisions, and later published a dataset of 1,944 manually annotated legal references from German court decisions \cite{r13}. While these efforts advanced retrieval and similarity modeling in the German legal domain, they did not provide a fully structured corpus that separates \textit{Tenor}, \textit{Tatbestand}, and \textit{Entscheidungsgründe} across all decisions.

In addition to datasets, other studies have highlighted how adding structural clarity to legal texts enables computational applications. Heckelmann \cite{r14} examined how legal agreements can be represented as executable smart contracts, thus defining a legal framework for their machine execution. While the focus is on fitting smart contracts into the existing framework of civil law, the motivation is the same: \textit{structure is a prerequisite for making legal data usable in downstream tasks}.

\section{Dataset}

We created our dataset from the official Open Legal Data dump (as of \textit{2022-10-18}\footnote{Open Legal Data dump. See \url{https://static.openlegaldata.io/dumps/de/2022-10-18/}}). While this raw dataset is available in JSONL format, the decision text is embedded within a \texttt{content} attribute with varying HTML structures. Our extraction pipeline focuses on separating three sections, normalizing court metadata, and extracting statutory and case references.

\subsection{Extraction Process}
We begin by parsing the HTML string in the \texttt{content} attribute with an HTML parser and iterate over visible elements limited to \texttt{p}, \texttt{h1}--\texttt{h4}, \texttt{td}, and a custom \texttt{rd} tag. The text is then normalized via whitespace collapsing, and empty and duplicate lines are skipped.

Next, we normalize the court metadata by resolving city and state identifiers using the public APIs from Open Legal Data\footnote{Open Legal Data API. \url{https://de.openlegaldata.io/api/}} (endpoints \texttt{/api/states/} and \texttt{/api/cities/}). Missing entries were set to \textit{Unspecified}.

Then, we perform section boundary detection by identifying specific, fixed-vocabulary headers. These section headers are recognized by two exact, line-level patterns per section, which are applied with case insensitivity and require a full-line match. The patterns are:

\begin{itemize}
    \item Compact form: \verb|^\s*<marker>\s*:*$|
    \item Spaced-letter form: \verb|^\s*<m a r k e r>\s*:*$|
\end{itemize}
The section vocabulary is fixed to \texttt{tenor}, \texttt{tatbestand}, \texttt{entscheidungsgründe}, and \texttt{gründe}\footnote{Included in the vocabulary as required for boundary detection, but later divided into \textit{Tatbestand} or \textit{Entscheidungsgründe}}. For each, we then test both the compact and space-letter variants (for example, \texttt{tenor} vs. \texttt{t e n o r}). The active section defaults to \texttt{tenor} until the first header is encountered and non-header lines are appended to the currently active section. This assumption is consistent with German legal drafting practice, where all court decisions typically begin with \textit{Tenor}.

German court decisions generally follow two drafting patterns. \textbf{\textit{Urteile}} (\textit{decisions with a hearing}) provide all three sections explicitly: \textit{Tenor}, \textit{Tatbestand}, and \textit{Entscheidungsgründe}. \textbf{\textit{Beschlüsse}} (\textit{decisions without a hearing}), although beginning with \textit{Tenor}, usually contain a \textit{Gründe} section subdivided by Roman numerals, where, \textit{Gründe I} corresponds to the \textit{Tatbestand} and \textit{Gründe II} to the \textit{Entscheidungsgründe}. When no subdivision is present, the entire \textit{Gründe} section is treated as \textit{Entscheidungsgründe}. This logic is consistently applied throughout our extraction pipeline.

Additionally, we also identify and extract the \textit{Rechtsmittelbelehrung}, which provides procedural instructions on available appeals. Although it is part of the published decision, it is not considered a substantive section of judicial reasoning; therefore, we store it separately in the schema.

Following section segmentation, all collected lines are processed with the Legal Reference Extraction tool\footnote{\href{https://github.com/openlegaldata/legal-reference-extraction}{https://github.com/openlegaldata/legal-reference-extraction}},
 which identifies legal citations and categorizes them by type (\texttt{law}, \texttt{case}). Each court decision is then stored as a single JSON object containing normalized metadata, sectioned text, and extracted references, allowing for direct use without additional preprocessing. A detailed JSON example entry showing this structure is available online\footnote{Example entry: \href{https://huggingface.co/datasets/harshildarji/openlegaldata\#example-entry}{https://huggingface.co/datasets/harshildarji/openlegaldata\#example-entry}}.

\subsection{Verification Process}\label{sec:verification}

Automatic section segmentation of decision texts can introduce subtle errors, such as misplaced boundaries or partial overlap between sections. To evaluate the reliability of our extraction process, we estimated the necessary sample size for manual verification using Cochran's formula with finite population correction. This approach ensures a statistically representative sample size for categorical evaluations, where each decision is either correctly or incorrectly segmented.

The initial sample size for an infinite population is given by:
\begin{equation}
    n_0 = \frac{Z^2 \cdot p \cdot (1 - p)}{e^2}
    = \frac{1.96^2 \cdot 0.5 \cdot 0.5}{0.05^2}
    \approx 384.16,
\end{equation}
\noindent\textit{where} $Z$ is the critical value for the chosen confidence level ($Z{=}1.96$ for $95\%$), $p$ is the estimated proportion of correct extractions (set to $0.5$ to maximize variance and yield the most conservative---i.e., largest---sample size estimate), and 
$e$ is the margin of error ($0.05$ for $5\%$).
Applying the finite population correction for $N{=}251{,}038$ decisions yields
\begin{equation}
    n = \frac{n_0}{1 + \frac{n_0 - 1}{N}}
    \approx 383.58.
    \label{eq:cochran}
\end{equation}

\noindent\textit{where} $N$ in this context serves as the population size. 
In practice, we use the conservative rounded sample size of $\mathbf{n=384}$.

Thus, 384 cases were selected uniformly at random and manually reviewed. 
Each sampled case was checked to confirm that the three sections (\textit{Tenor}, \textit{Tatbestand}, and \textit{Entscheidungsgründe}) were correctly identified, with correctness defined strictly as the absence of overlap between sections. 
The manual review confirmed correct segmentation in 97.40\% of cases, with errors mainly due to rare formatting irregularities in the HTML. 
Based on this result, we computed a 95\% confidence interval using the normal approximation with finite population correction, which yields $(0.9581,\;0.9899)$ for the full dataset of 251,038 decisions. 
This confirms that the true proportion of correctly segmented cases lies between 95.8\% and 98.9\%, indicating a high level of reliability for the extraction process.

\subsection{Section Coverage}
Table~\ref{tab:section-coverage} shows the coverage of three sections within our dataset. The \textit{Tenor} is present in 87.7\% of cases, while the \textit{Tatbestand} appears in 65.4\%, and the \textit{Entscheidungsgründe} in 95.1\%. The difference in coverage of \textit{Tatbestand} and \textit{Entscheidungsgründe} reflects differences in drafting practice between decision types, with \textit{Urteile} usually containing both sections explicitly, while in \textit{Beschlüsse}, the factual background is sometimes merged into the reasoning when no subdivision into \textit{Gründe I} and \textit{Gründe II} is provided. The \textit{Rechtsmittelbelehrung}, which provides procedural instructions on available appeals, appears in 8,335 decisions (3.32\% of the corpus).

\begin{table}[H]
\centering
\begin{minipage}{0.45\linewidth}
\centering
\caption{Section coverage over 251{,}038 decisions.}
\begin{tabular}{lr}
\toprule
Section & Count \\
\midrule
Tenor & 220{,}273 (87.7\%) \\
Tatbestand & 164{,}222 (65.4\%) \\
Entscheidungsgründe & 238{,}666 (95.1\%) \\
\bottomrule
\end{tabular}
\label{tab:section-coverage}
\end{minipage}\hfill
\begin{minipage}{0.45\linewidth}
\centering
\caption{Structural composition of decisions.}
\begin{tabular}{lr}
\toprule
Structure & Count \\
\midrule
All three sections & 144{,}383 (57.5\%) \\
Only Tenor + Ent. & 63{,}720 (25.4\%) \\
Only Tenor & 11{,}388 (\hphantom{0}4.5\%) \\
\bottomrule
\end{tabular}
\label{tab:struct-comp}
\end{minipage}
\end{table}

In addition, there are \textbf{176} decisions (\textbf{0.07\%} of the total) in which all sections are absent. These correspond to cases where the original \texttt{content} field is blank. Table~\ref{tab:struct-comp} shows how the different sections are combined within the dataset. A majority, 57.5\%, of the decisions contain all three sections, while only 25.4\% of decisions contain only the \textit{Tenor} and \textit{Entscheidungsgründe}, and a small number of decisions, 4.5\%, contain only the \textit{Tenor} section.

\section{Conclusion and Future Work}
In this paper, we presented a dataset of 251,038 German court decisions derived from the Open Legal Data dataset. Our dataset provides a consistent structure by segmenting decisions into \textit{Tenor}, \textit{Tatbestand}, and \textit{Entscheidungsgründe}, addressing the inconsistent formatting and incomplete markers of the original raw HTML. We also evaluated the extraction pipeline using a statistically representative random sample, using Cochran's formula. The manual verification confirmed the reliable separation in approximately 97\% of cases. The dataset is available in JSONL format and includes metadata as well as extracted references, making it directly usable for further research without additional preprocessing.

We are also currently using this dataset to build a RAG system for German legal texts. The segmented structure allows our retrieval pipeline to separately index case summaries, statutory references, and reasoning paragraphs, which are then aligned with user queries. Our future work will focus on improving retrieval quality by adding ranking and reranking strategies. We also plan to evaluate the RAG system across subtasks that further demonstrate the value of segmentation. These include statute retrieval, where reasoning passages must be aligned with cited provisions, reasoning coverage, where factual context and arguments need to be distinguished, and interpretability, where users benefit from seeing only the most relevant section of a decision. Finally, we will extend this corpus with additional court decisions as they become available and fine-tune the handling of legal drafting variations.

\bibliographystyle{vancouver} 
\bibliography{references}

\end{document}